\documentclass{Interspeech}

\interspeechcameraready

\title{WIND: Accelerated RNN-T Decoding with Windowed Inference for Non-blank Detection}

\author{Hainan}{Xu}
\author{Vladimir}{Bataev}
\author{Lilit}{Grigoryan}
\author{Boris}{Ginsburg}

\affiliation{}{NVIDIA Corporation}{U.S.A.}
{\email{hainanx@nvidia.com}}

\keywords{speech recognition, RNN-T, RNN-Transducers, parallel computing}

\usepackage{comment}
\usepackage{multirow}
\usepackage[noend]{algorithmic}
\usepackage{algorithm}
\usepackage{graphicx}

\begin{document}

\maketitle

% the abstract here must exactly match the abstract entered into the paper submission system
\begin{abstract}

%RNN Transducer (RNN-T) has become a popular choice for automatic speech recognition due to its simple model architecture and competitive performance, however, its inference efficiency is limited by the sequential nature of frame processing, particularly when handling blank symbol predictions. 

We propose Windowed Inference for Non-blank Detection (WIND), a novel strategy that significantly accelerates RNN-T inference without compromising model accuracy. During model inference, instead of processing frames sequentially, WIND processes multiple frames simultaneously within a window in parallel, allowing the model to quickly locate non-blank predictions during decoding, resulting in significant speed-ups. We implement WIND for greedy decoding, batched greedy decoding with label-looping techniques, and also propose a novel beam-search decoding method. Experiments on multiple datasets with different conditions show that our method, when operating in greedy modes, speeds up as much as 2.4X compared to the baseline sequential approach while maintaining identical Word Error Rate (WER) performance. Our beam-search algorithm achieves slightly better accuracy than alternative methods, with significantly improved speed. We will open-source our WIND implementation.
    
\end{abstract}

\section{Introduction}

In the recent decade or so, End-to-end automatic speech recognition (ASR) has seen remarkable progress, with RNN-T \cite{graves2012sequence} emerging as one of the dominant architectures for speech applications.
Various open-source toolkits offer high-quality implementations for RNN-T models, including ESPnet \cite{watanabe2018espnet}, SpeechBrain \cite{ravanelli2021speechbrain}, and NeMo \cite{kuchaiev2019nemo} etc.
Compared to alternative methods like CTC \cite{graves2006connectionist}, RNN-T achieves improved accuracy at increased computational cost. As deployment scenarios become more diverse and demanding, there is an increasing need to improve RNN-T's efficiency without compromising its recognition accuracy.
A significant computational bottleneck in RNN-T inference stems from its frame-by-frame processing mechanism during decoding, where the model loops over all input frames, with majority of decoding steps predicting blank tokens. Blank predictions of RNN-T models are time-consuming but do not contribute to the final decoding outputs, marking significant waste.

In this paper, we introduce \emph{Windowed Inference for Non-blank Detection} (WIND), a novel parallel strategy that accelerates RNN-T inference. Instead of processing frames sequentially, WIND processes multiple frames simultaneously within a configurable window, allowing the model to locate the next non-blank symbol efficiently, significantly improving the inference speed of RNN-T models with minimal memory overhead.
%At the time of submission, WIND supports efficient greedy decoding, label-looping \cite{bataev2024label} batched greedy decoding. We also propose a novel beam search algorithm using the WIND strategy. 
The key contributions of this work are:

\begin{enumerate}
    \item We propose the WIND strategy for RNN-T inference, which uses parallelized computation on a window of frames to detect non-blanks quickly.
    \item Greedy (batch=1) WIND achieves up to 2.40X inference speed-up compared to the standard algorithm.
    \item Batched greedy WIND achieves up to 1.26X speed-up to highly optimized label-looping methods with CUDA graphs.
    \item Beam-search WIND archives significantly better speed-accuracy tradeoff than alternative methods like Alignment-length synchronous decoding (ALSD)~\cite{saon2020alignment} and Modified adaptive expansion search (MAES) ~\cite{kim2020accelerating}. 
    \item WIND can combine with alternative methods like CUDA-graphs to further optimize RNN-T inference.
    \item All evaluations in this paper use publicly accessible checkpoints and public data. We will open-source our WIND implementations so that the results can be fully reproduced.

\end{enumerate}

\section{Related Work}

Considerable work has investigated speeding up the inference of Transducer models.
One way to do that is through improvement in the model architecture. For example, \cite{Ghodsi2020stateless} replaced the RNN components of the model decoder with a stateless network, delivering consistent model speedup with small performance degradation. 
\cite{xu2022multi} proposed a multi-blank Transducer, where blank tokens can cover multiple frames, and therefore reducing the number of decoding steps;
\cite{xu2023efficient} proposed a Token-and-Duration Transducer (TDT), which jointly predicts labels and durations that the label covers, further reducing the number of decoding steps. 
\cite{xu2024three} proposed a stochastic masking mechanism during Transducer model training, so that the model can also support non-autoregressive decoding without running the decoder. \cite{yang2023blank} proposed a way to jointly train CTC and RNN-T model, and use CTC's blank prediction to run pruning on the RNN-T decoding, achieving speedup during inference.

Another direction is to improve the RNN-T efficiency through algorithmic improvement during inference. CUDA Graphs \cite{galvez2024speed} improves the efficiency of Transducer inference by reducing the CUDA launch time. Label-looping algorithm \cite{bataev2024label} synchronizes decoder operations during batched inference and brings significant speedups. Regarding beam-search for RNN-T models, in addition to the original algorithm proposed in \cite{graves2012sequence}, some more efficient variants, e.g. alignment-length synchronous decoding \cite{saon2020alignment}, modified adaptive-expansion search \cite{kim2020accelerating}, one-step constrained search \cite{kim2020accelerating2}, token-wise beam search \cite{keren2023token} etc, achieve faster inference by employing different model assumptions to reduce the search space.

%Our inference algorithm works in similar ways with \cite{xu2022multi} and \cite{xu2023efficient}, without having to retrain the model to explicitly support duration prediction from models. 

\section{Method}

\subsection{Background: RNN-T}
\begin{algorithm}
{\fontsize{8.1}{9.7}\selectfont 

   \caption{RNN-T Inference Algorithm}
   \label{RNNT_algo}
\begin{algorithmic}[1]
   \STATE {\bfseries input:} encoder output enc [T, D]
    \STATE hyp = []
    \STATE t = 0
    \WHILE{$t <$ T}
    \STATE dec = decoder(hyp)
    \STATE joined = joiner(enc[$t$,:], dec[-1])
%    \STATE token\_probs = softmax(joined)
    \STATE label = argmax(joined)
    \IF{label is not blank}
    \STATE hyp.append(label)
    \ELSE
    \STATE $t$ += 1 
    \ENDIF
    \ENDWHILE
    \STATE {\bfseries return} hyp 
\end{algorithmic}
}
\end{algorithm}

% \begin{figure}
%     \centering
%     \includegraphics[width=0.7\linewidth]{rnnt_arch.png}
%     \caption{Model architecture of RNN-T models.}
%     \label{fig:rnnt_arch}
% \end{figure}

%Before introducing our WIND algorithm, we first provide the background information on RNN-T models. 
%Figure \ref{fig:rnnt_arch} shows the architecture of 
An RNN-T model consists of an encoder, a decoder, and a joiner. The encoder and decoder extract high-level information from acoustic frames and textual history, respectively, and the joiner combines the information from the two components to generate a probability distribution over the vocabulary. The vocabulary of an RNN-T model can be words, graphemes, or subwords (e.g. byte-pair encoding \cite{sennrich2015neural,kudo2018sentencepiece}), and must include a special ``blank'' token. The interaction of tokens in the vocabulary and the inference process is as follows,
\begin{enumerate}
    \item when a non-blank symbol is predicted, the symbol is added to the output; meanwhile, the model stays at the same acoustic frame $t$ for the next decoding step, and updates its decoder representation by feeding the symbol to its decoder; 
    \item when a blank symbol is predicted, the model keeps its decoder representation unchanged, and increments $t$ by one to access the next acoustic frame.
\end{enumerate}

The standard inference procedure for RNN-T models is shown in Algorithm \ref{RNNT_algo}. The algorithm scans the encoder output from left to right in a frame-by-frame fashion. For any frame, it uses the joint network to combine its information with the decoder output computed from the current partial hypothesis and generate a probability distribution over vocabulary.

\subsection{Greedy Inference with WIND Models}

\begin{algorithm}
{\fontsize{8.1}{9.7}\selectfont 
   \caption{WIND Algorithm}
   \label{WIND_algo}
\begin{algorithmic}[1]
   \STATE {\bfseries input:} encoder output enc [T, D]
    \STATE hyp = []
    \STATE t = 0
    \WHILE{$t <$ T}
    \STATE dec = decoder(hyp)
    \STATE n = min(window-size, len(enc) - t) 
    \STATE joined = joiner(enc[$t$:$t+n$,:], dec[-1]) \# [n, V]
%    \STATE token\_probs = softmax(joined)
    \STATE labels = argmax(joined, dim=-1) \# [n]
    \IF{labels are all blanks}
    \STATE $t$ += n
    \ELSE
    \STATE i = smallest index so that labels[i] != blank
    \STATE hyp.append(labels[i])
    \STATE $t$ += $i$
    \ENDIF
    \ENDWHILE
    \STATE {\bfseries return} hyp 
\end{algorithmic}
}
\end{algorithm}

We notice that in Algorithm \ref{RNNT_algo},  the decoder representation would only change after a non-blank symbol is predicted; if consecutive blanks are predicted, all those steps reuse the same decoder representation. %, and there is no computational dependency among them.
We recognize this as a potential for efficiency improvements, and propose to parallelize the computation of those frames with Algorithm \ref{WIND_algo}. At any decoding step, we use the joiner to combine the information of current decoder representation, with multiple consecutive acoustic frames within a window. This allows us to locate non-blank frames faster, reducing the decoding time. The algorithm can be easily applied to batched inference, including the more efficient label-looping batched algorithms \cite{bataev2024label}. Due to space constraints, we omit the algorithm description and refer the readers to our open-source implementation for details.

% \subsection{Window-based Greedy Inference}

% In the previous subsection, the ``greedy'' algorithm finds the arg-max token per frame. While this heuristic is efficient and usually works well, it might miss tokens that might accumulate to better probabilities by summing over multiple frames. As an improvement,
% we propose a window-based greedy search algorithm, which aims to find the best prediction based on information within the whole window.

% With this probability defined, we introduce the window-based greedy algorithm in Algorithm \ref{WIND_window_algo}.

% \begin{algorithm}[h]
%    \caption{Window-based Greedy Algorithm}
%    \label{WIND_window_algo}
% \begin{algorithmic}[1]
%    \STATE {\bfseries input:} encoder output enc
%     \STATE hyp, t = [], 0

%     \WHILE{$t <$ len(enc)}
%     \STATE dec = decoder(hyp)
%     \STATE w = min(window\_size, len(enc) - t) 
%     \STATE joined = joiner(enc[$t$:$t+w$,:], dec) \# [w, V]
%     \STATE compute P'(.) \# the P'(v, t) we introduced
%     \STATE label, jump = argmax(P'(.)) \# both are integers
%     \IF{label is blank}
%     \STATE $t$ += n \# jump out of the window
%     \ELSE
%     \STATE hyp.append(label)
%     \STATE $t$ = jump
%     \ENDIF
%     \ENDWHILE
%     \STATE {\bfseries return} hyp 
% \end{algorithmic}
% \end{algorithm}

%\subsection{Label-looping Batched Greedy Inference}

\subsection{A Novel Beam-search  Algorithm for WIND}

\begin{algorithm}
{\fontsize{8.1}{9.7}\selectfont 
   \caption{WIND Beam-search Algorithm}
   \label{WIND_beam_algo}
   % \small
\begin{algorithmic}[1]
   \STATE {\bfseries input:} encoder output enc [T, D], window  $N$, beam  $K$
    \STATE t2hyps = \{0: [Hyp(tokens=[], score=0.0)]\}
    \WHILE{t2hyps is not empty}
        \STATE t = min(t2hyps.keys())
        \IF{$t == $ T}
            \STATE break
        \ENDIF
        \STATE hyps = t2hyps.pop(t)  \# read and remove
%        \STATE hyps = prefix\_search(hyps)
        \STATE hyps = recombine\_prune\_prefix\_search(hyps)
        \STATE n = min(N, T - t) \# determine the window size
        \STATE windows = enc[t:t+n,:] \# [n, D]
        \STATE joined = joiner(windows, dec\_state(hyps)) \# [B, n, V]
        \STATE compute $P_b'(v, t)$ for all V, t, and b
        \STATE labels, jumps = top\_expandions\_for\_each\_b($P_b'(v, t)$, K) \# both of shape [B, K]
        \FOR{b in range(B)}
            \FOR{k in range(K)}
                \STATE new\_hyp = hyps[b] \# take existing hyp
                \STATE new\_token, jump = labels[b][k], jumps[b][k]
                \IF{new\_token is not blank}
                    \STATE new\_hyp.tokens.append(new\_token)
                    \STATE new\_hyp.scores += log($P_b$(new\_token, jump))
                    \STATE t2hyps[t + jump].append(new\_hyp)
                \ELSE
                    \STATE new\_hyp.scores += log($P_b$(blank, n-1))
                    \STATE t2hyps[t + n].append(new\_hyp)
                \ENDIF
            \ENDFOR
        \ENDFOR        
    \ENDWHILE
    \STATE {\bfseries return} best\_hyp\_in(t2hyps[T])
\end{algorithmic}
}
\end{algorithm}

%Before introducing our beam-search Algorithm, let's first define some terms. 
Given a vocabulary set $\mathcal{V}$ (including blank, represented as $\emptyset$), a window size $w$, and 
say we are processing the acoustic frame $t_0$.
Based on the current decoder state, the RNN-T model computes a probability distribution of $P(v |t_0 + t)$ for $v \in \mathcal{V}$ and $t \in \{0, 1, ..., w - 1\}$. Now we define a new probability distribution of $P'(v, t)$ as the ``probability of predicting $v$ as the first output in the window at time $t_0 + t$''. This probability is interpreted as,
\begin{enumerate}
    \item for $v = \emptyset$, it must be predicted at the last frame of the window, and the probability is computed as product of probability of predicting blanks at \emph{all} frames in the window;
    \item for $v \neq \emptyset$, it is the probability of emitting all blanks for frames $t_0, t_0 + 1, ..., t_0 + t - 1$, and then emitting $v$ at  $t_0 + t$.
 \end{enumerate}

With this definition, $P'(v, t)$ can be computed as,
 
%    \scalebox{1.0}{
\begin{equation} 
P'(v, t) = \begin{cases}
        0 & v = \emptyset, t < w - 1
        \\
        \displaystyle \prod_{t'=t_0}^{t_0 + w - 1} P(\emptyset | t') & v = \emptyset, t = w - 1
        \\
        P(v | t_0 + t) \displaystyle \prod_{t'=t_0}^{t_0 + t-1} P(\emptyset | t'), & v \neq \emptyset
        \end{cases}
\end{equation}
%}

With $P'(v, t)$ defined, we present our  WIND-based beam-search for in Algorithm \ref{WIND_beam_algo}\footnote{For simplicity, we omit some details regarding safeguard code to avoid the loop from staying at the same time-stamp indefinitely. Readers are referred to our open-sourced implementation for such details.}.
We represent a hypothesis with a tuple of (List(int), float), where the list is the sequence of tokens and float is the score. The algorithm maintains a t2hyps map, storing partial hypotheses that end at a specific time step. During processing, it takes the hypotheses associated with the smallest time steps $t$, processes them (more on this later). Then, for each of the hypotheses we compute the top-beam expansions in the window enc[t:t+n,:]. Note that at line 13, the topk function selects the best candidates from all $|\mathcal{V}| \times n$ possible combinations using $P'(v, t)$, allowing the algorithm to jump multiple frames during the search. 
For each of the chosen expansions, it updates the hypotheses and adds them t2hyps at the correct timestep for later processing. Once the Algorithm reaches the end, it returns the hypothesis with the best score.\footnote{Note, our proposed algorithm shares similarities with \emph{token-wise beam-search} \cite{keren2023token} in using a window (chunk) of frames as the unit of computation; other than that, the decoding algorithms of the two methods are very different. In particular, a major difference is that \cite{keren2023token} aggregates the emissions from chunks, while in our algorithm we still attribute emissions to a specific frame within the window, so we can still generate accurate time-stamps from the beam-search. } 

The  {``recombine\_prune\_prefix\_search''} function at line 8 performs the following steps,
\begin{enumerate}
    \item it checks if there are duplicate hypotheses, and if so, combines them into one with their probabilities summed.
    \item for all hypothesis pairs (A, B) where A is a prefix of B, it computes the probability of ``completing A into B'' by emitting extra symbols at the current frame; it then removes A and adds the extra probability mass to B. Note that this function was originally proposed in Section II-B and lines 5-7 of Algorithm 1 of \cite{kim2020accelerating}, where readers can find more details.
    \item it keeps top beam hyps in the set.
\end{enumerate}

\section{Experiments}

We conduct our experiments using Conformer-RNNT \cite{gulati2020conformer,rekesh2023fast} implementation from the NeMo \cite{kuchaiev2019nemo} toolkit. 
All models are public checkpoints built on 80-dimensional filter bank features extracted at 25ms frames with 10ms strides, and BPE of size 1024 as text representation.
%We list their download pages where readers can find hyperparameter details. 
To measure inference time for all experiments, we first do a ``warm-up'' run, and then run three consecutive decoding runs and take the average time for those three runs with two A6000 GPUs.

\begin{table}[t]
    \centering
    \caption{Librispeech Word-Error-Rates (WER\%) and decoding time (seconds) of Parakeet-RNNT-1.1b and -0.6b models at different window-sizes. Window-size=1 is the standard (baseline) algorithm against which relative speedup is measured.}
    \begin{tabular}{c c c c c}
    \toprule
 model                               & window        & WER \%& time (s)  & rel. speed-up\\
 \midrule
 \multirow{5}{*}{1.1b}              & 1             & 2.70  & 176        & - \\
                                    & 2             & 2.70  & 122        & 1.44X \\
                                    & 4             & 2.70  & 91        & 1.93X \\
                                    & 8             & 2.70  & 83        & 2.12X \\
                                    & 16            & 2.70  & 83        & 2.12X \\
                                    % & 32            & 2.70  & 82        & 2.15X \\
 \midrule
 \multirow{5}{*}{0.6b}              & 1             & 3.31  & 161        & -      \\
                                    & 2             & 3.31  & 105       & 1.53X \\
                                    & 4             & 3.31  & 74        & 2.18X \\
                                    & 8             & 3.31  & 67        & 2.40X \\
                                    & 16            & 3.31  & 67        & 2.40X \\
                                    % & 32            & 3.31  & 66        & 2.44X \\
                                    
    \bottomrule
    \end{tabular}
    \label{parakeet_0.6b}
\end{table}

\begin{table}
    \centering
    \caption{Parakeet models' decoding time (seconds) comparison between the original label-looping VS WIND version of label-looping batched greedy inference. Window-size=8 for the WIND algorithm. Evaluation is done on librispeech test-other.}
    \begin{tabular}{c c c c c}
    \toprule
 model                              & batch      & baseline & WIND      & rel. speed-up\\
 \midrule
 \multirow{4}{*}{1.1b}              & 2             & 142   & 120       & 1.18X \\
                                    & 4             & 91    & 77        & 1.18X \\
                                    & 8             & 66    & 57        & 1.16X \\
                                    & 16            & 54    & 48        & 1.13X \\
                                    % & 32            & 31    & 66        & 1.148X \\
\midrule
 \multirow{4}{*}{0.6b}              & 2             & 105   & 87        & 1.21X \\
                                    & 4             & 68    & 54        & 1.26X \\
                                    & 8             & 48    & 39        & 1.23X \\
                                    & 16            & 38    & 32        & 1.19X \\
                                    % & 32            & 32    & 28        & 1.093X \\
                                    % 
    \bottomrule
    \end{tabular}
    \label{batched_greedy}
\end{table}

\subsection{Greedy Decoding}

% In Table  \ref{parakeet_0.6b}, we compare WIND with standard RNN-T inference algorithms in greedy modes. We use Parakeet-RNNT-0.6b and 1.1b models ({\url{hf.co/nvidia/parakeet-rnnt-0.6b} and \url{hf.co/nvidia/parakeet-rnnt-1.1b}}). 
In Table~\ref{parakeet_0.6b}, we compare WIND with standard RNN-T inference algorithms in greedy modes. We use Parakeet-RNNT-0.6b~\footnote{\url{hf.co/nvidia/parakeet-rnnt-0.6b}} and 1.1b~\footnote{\url{hf.co/nvidia/parakeet-rnnt-1.1b}} models. 
We report decoding results with different window-sizes, where window-size=1 represents the standard RNN-T algorithm. As we can see, in all settings the WIND algorithm achieves identical WER compared to the standard algorithm, at significantly improved speed. Larger window-size would bring larger speedups although the effect plateaus for window-size 8 and up.

\subsection{Batched Greedy Results}

The results of batched greedy search are shown in Table \ref{batched_greedy}. Note, we omit the WER numbers since they are identical. The relative speedup is computed as the ratio between time of the standard algorithm and the WIND algorithm. Both algorithms use the label-looping \cite{bataev2024label} methods.
We can see that WIND algorithm brings consistent speedup to the standard algorithm, with the effect more pronounced for relatively smaller batches.

\begin{table}
    \centering
    \caption{Parakeet-RNNT-0.6b's WER and decoding time (seconds) of different beam-search algorithms for English Librispeech test-other (up) and Slurp test (down).}
        \addtolength{\tabcolsep}{-0.3em}
    \begin{tabular}{c c c c c c c c c }
    \toprule
           & \multicolumn{2}{c}{greedy}   & \multicolumn{2}{c}{beam=2} & \multicolumn{2}{c}{beam=3}    & \multicolumn{2}{c}{beam=4}  \\
   & WER & time    & WER          & time               &  WER          & time          &  WER          & time          \\
\midrule
alsd   & \multirow{3}{*}{3.31} &  \multirow{3}{*}{161}      &  3.24 & 297 & 3.24 & 384  & 3.25 & 470 \\
maes   &   &                                                &  3.29 & 169 & 3.29 & 185  & 3.28 & 204  \\
beam   &   &                                                &  3.26 & 388 & 3.26 & 600  & 3.26 & 813  \\
%tsd    &   &                                                &  3.26 & 1903  \\
\midrule
wind   & 3.31 & 67                                          & 3.25 & 105 & 3.26 & 114  & \textbf{3.23} & 120  \\
\midrule
\midrule
alsd  & \multirow{2}{*}{17.93 }  & \multirow{2}{*}{319 }            &  17.77 & 622 & 17.74 & 774 & 17.82 & 949 \\
maes     &&                                                 &  17.83 & 357 & \textbf{17.72} & 394 & 17.73 & 435  \\
\midrule
wind     &  17.93 &  139                                    &  17.80 & 221 & \textbf{17.72} & 254 & 17.79 & 283 \\
% wind (4)    &   \\
% wind (2)    &   \\
                                    
    \bottomrule
    \end{tabular}
    \label{beam_slurp}
\end{table}

\begin{table}
    \centering
    \caption{German ASR model \url{stt_de_conformer_transducer_large}'s WER and decoding time (seconds) of different beam-search algorithms on German Voxpopuli (up) and Multilingual Librispeech (down).}
        \addtolength{\tabcolsep}{-0.3em}
    \begin{tabular}{c c c c c c c c c}
    \toprule
    &    \multicolumn{2}{c}{greedy}     & \multicolumn{2}{c}{beam=2}        & \multicolumn{2}{c}{beam=3}    & \multicolumn{2}{c}{beam=4}  \\
    & WER          & time               &  WER          & time  & WER          & time    &  WER          & time      \\
\midrule
alsd &  \multirow{2}{*}{8.85} & \multirow{2}{*}{130}    & 8.76  & 606   & 8.63  & 719   & 8.63  & 871 \\
maes  &   &                                             & 8.69  & 217   & 8.69  & 254   & \textbf{8.64}  & 281   \\
\midrule
wind  &  8.85 &  61                                     & 8.72  & 109   & \textbf{8.64}  & 124   & \textbf{8.64}  & 135   \\
\midrule
\midrule
alsd &  \multirow{2}{*}{3.85} & \multirow{2}{*}{437}    & 3.85  & 1840  & 3.80  & 2127  & 3.82  & 2556 \\
maes  &   &                                             & 3.87  & 620   & 3.83  &  718  & 3.80  & 780    \\
\midrule
wind  & 3.85  & 167                                     & 3.85  & 302   & \textbf{3.78}  & 351   & 3.81  & 382   \\

    \bottomrule
    \end{tabular}
    \label{beam_mls}
\end{table}

\subsection{Beam Search Results}

Since beam-search algorithms can change the WER of different models, to provide a better picture of how WIND beam-search compares to alternative methods, we report our beam-search results on multiple datasets across languages, including Voxpopuli and Multilingual Librispeech for German, and Librispeech test-other and Slurp for English. The German evaluation also uses a publicly accessible German model \footnote{ \url{hf.co/nvidia/stt_de_conformer_transducer_large}}.
Tables \ref{beam_slurp} and \ref{beam_mls} show the results where we compare with the original beam-search (beam)  \cite{graves2012sequence}, \emph{alignment-length synchronous decoding} (alsd)  \cite{saon2020alignment} and \emph{modified adaptive expansion search} (maes)  \cite{kim2020accelerating}. We include ``greedy'' decoding results as well for context.

We can see the WIND beam-search achieves better speed-accuracy tradeoff than all alternative methods. Since the Librispeech test-other numbers show that ``beam'' takes significantly more time to run without performance gains, we don't include it for other experiments. Notably, WIND with beam=4 runs faster than all alternative methods with beam=2; in 3 of the 4 cases, it's faster than (Non-WIND) greedy decoding. We also see that for all datasets, WIND can achieve the best accuracy (\textbf{bold} numbers) among all methods, and in terms of a tie, WIND uses significantly less time than alternative methods.

\section{Analysis}

\subsection{Memory Footprint}

Due to the decoding window used by the WIND algorithm, more GPU memory usage of the WIND algorithm is expected. However, our empirical studies reveal that the WIND algorithm does not use noticeably more memory. Note, our models consist hundreds of millions of parameters, which require Gigabytes of GPU memory for storing model weights and hidden activations;
the WIND algorithm only increases memory usage by [window-size $\times$ vocabulary-size]. In our experiments, this is at most $16 \times 1025 = 16400 $ floats, which amounts to around 63 Kb of memory if using float32 datatypes. Therefore, this added memory overhead is negligible.

\subsection{Jump Distribution}

\begin{figure}
    \centering
    \caption{Jump distributions of WIND inference with Parakeet-RNNT-1.1b model on Librispeech test-other. Bars are ordered so that smaller jumps are on the left, and the larger right, so readers viewing this on black-and-white printed paper can infer which bar represents which jump interval without using color information.}
    \includegraphics[width=0.85\linewidth]{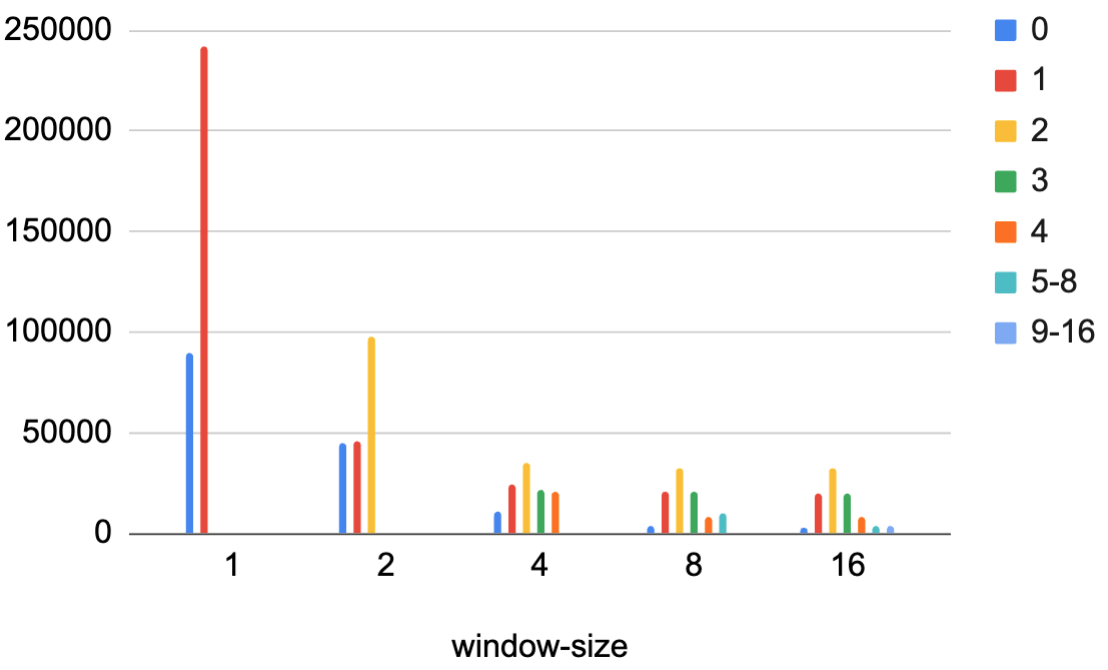}

    \label{frequency}
\end{figure}

We plot the distributions of different jump sizes for WIND inference with different window sizes, as shown in Figure \ref{frequency}. We see that with window-size 1 and 2, a significant probability mass is allocated to the longest possible jump; once the window-size becomes larger, the inference would still use them as needed but the the long jumps are relatively less used. The overall distribution for window-sizes 8 and 16 are not very different, which explains their similar decoding speed previously shown in Tables \ref{parakeet_0.6b} for those larger window-sizes.

\subsection{Incorporation wth CUDA-graphs}
CUDA-graph \cite{galvez2024speed} is a technique to speedup GPU processing by reducing GPU kernel launch time. We  implement CUDA-graph optimization for label-looping batched WIND inference, and show their comparison in Table \ref{cuda_graph}. 
To provide more context, we include results for a comparable TDT~\cite{xu2023efficient} public checkpoint~\footnote{\url{hf.co/nvidia/parakeet-tdt-1.1b}}. We can see that while CUDA-graph optimization brings significant speedup for inference in all cases, WIND brings consistent further speed improvement with or without CUDA-graphs, especially when we compare decoder-only time. 
We also see that WIND inference with RNN-T is faster than TDT, indicating its strength in improving Transducer model efficiency without the need to retrain a model with a modified loss function.

\begin{table}[]
    \centering
    \caption{Decoding time of different decoding algorithms for batched greedy search, with Parakeet 1.1b RNN-T and TDT models, with and without CUDA-graph. Batch=16 for all runs. The WIND inference uses a window size of 8. ``Decoder-only time'' refers to the time used for inference excluding the encoder computation.}
    \begin{tabular}{c c c}
    \toprule
       model-decoding   & total-time & decoder-only time  \\
    \midrule
  RNN-T                 & 53.91      & 12.45 \\
  + cuda-graph          & 45.94      & 4.23 \\
    \midrule
  RNN-T WIND            & 48.05      & 6.45 \\
  + cuda-graph          & 43.85      & 2.10 \\
    \midrule
  TDT                   & 48.49      & 6.93 \\
  + cuda-graph          & 44.00      & 2.14 \\
\bottomrule
    \end{tabular}

    \label{cuda_graph}
\end{table}

\section{Conclusion and Future Work}
%We will work on adding CUDA graph support to our methods. 
In this paper, we propose an Windowed Inference
for Non-blank Detection (WIND) strategy, which helps improve decoding speed of RNN-T for both greedy, batched greedy inference; we also propose a novel WIND beam-search method that achieves better speed-accuracy tradeoff than alternative beam-search methods.
In the future, we will also develop a more efficient WIND algorithm that combines beam search and batching.

\bibliographystyle{IEEEtran}
\bibliography{mybib}

\end{document}